\documentclass[letterpaper, 10 pt, conference]{ieeeconf}

\IEEEoverridecommandlockouts

\usepackage[utf8]{inputenc}
\usepackage[T1]{fontenc}
\usepackage{microtype}

\usepackage{graphicx}
\usepackage{algorithm}
\usepackage{algpseudocode}
\usepackage{subcaption}
\usepackage{lipsum}
\usepackage{enumerate}
\usepackage[hidelinks]{hyperref}

\usepackage{amsmath}
\usepackage{amssymb}
\usepackage{fontawesome5}

\usepackage{xspace}
\usepackage{csquotes}

\usepackage{todonotes}

\usepackage{tikz}
\usetikzlibrary{calc, arrows, arrows.meta, snakes, shapes}
\pgfdeclarelayer{bg}    
\pgfsetlayers{bg,main}  

\newtheorem{definition}{Definition}
\newtheorem{example}{Example}

\usepackage{cleveref}
\crefname{figure}{Fig.}{figures}
\Crefname{figure}{Figure}{Figures}
\crefname{definition}{Def.}{definitions}
\Crefname{definition}{Definition}{Definitions}
\crefname{section}{Sect.}{sections}
\Crefname{section}{Section}{Sections}
\crefname{table}{Table}{tables}
\Crefname{table}{Table}{Tables}

\newcommand{\uParam}{\ensuremath{\gamma}\xspace}

\newcommand{\probMC}{\textnormal{\textbf{P}}\xspace}

\newcommand{\rew}{\textnormal{R}\xspace}
\newcommand{\policy}{\ensuremath{\pi}\xspace}
\newcommand{\MDP}{\ensuremath{\mathcal{M}}\xspace}
\newcommand{\MC}{\ensuremath{\mathcal{M}}\xspace}

\newcommand{\pathMC}{\historyMC}
\newcommand{\historyMC}{\ensuremath{h}\xspace}

\newcommand{\estimatedDis}{\ensuremath{D}\xspace}

\newcommand{\E}[1]{\mathbb{E}[#1]}

\definecolor{mainblue}{HTML}{254E70}
\definecolor{mainred}{HTML}{C1666B}
\definecolor{mainyellow}{HTML}{E9D985}
\definecolor{maingreen}{HTML}{2C5530}

\tikzstyle{textNode} = [
anchor=center,
]
\tikzstyle{component} = [
draw=black,
fill=white,
text opacity=1,
textNode,
rounded corners,
minimum height=16pt,
minimum width=20pt,
line width=0.75
]
\tikzstyle{wideComponent} = [
component,
minimum width=3.5cm
]

\tikzstyle{tallComponent}= [
component,
minimum height=1.5cm
]

\tikzstyle{bigCompontent} = [
component,
minimum height=1.5cm,
minimum width=2.2cm
]

\tikzstyle{operationFlow} = [
line width=0.75pt,
-Latex
]
\tikzstyle{round node} = [circle, fill=white, draw=black, text=black, line width=0.75pt]
\tikzstyle{state} = [round node, draw=black, shape=ellipse, style={minimum width=0.75cm}, inner sep = 0pt, style={minimum height=0.75cm},]
\tikzstyle{accepting} = [double]
\tikzstyle{arrow} = [line width=0.75pt,
-Latex]

\tikzstyle{mdpLabel} = [
sloped
]

\tikzstyle{noRisk} = [draw=mainblue, text=mainblue]
\tikzstyle{risk} = [draw=mainred, text=mainred]

\tikzstyle{background} = [draw=mainblue, fill=cyan, fill opacity=0.025]

\newcommand{\oilDrop}{\text{\textcolor{black}{\faTint}} } 
\newcommand{\goalFlag}{\text{\textcolor{black}{\faFlagCheckered}} }

\usepackage{tikz}
\usepackage{pgfplots}

\title{\LARGE\bfseries Risk-Averse Planning and Plan Assessment for Marine Robots\,*}

\author{Mahya Mohammadi Kashani$^{1}$, Tobias John$^{2}$,
  Jeremy P.\:Coffelt$^{3}$, Einar Broch Johnsen$^{2}$ and Andrzej W\k asowski$^{1}$
\thanks{*This project has received funding from the European Union's Horizon 2020 research and innovation programme under the Marie Sk\l{}odowska-Curie grant agreement No 956200 REMARO.}
\thanks{$^{1}$Mahya Mohammadi Kashani and Andrzej Wąsowski are with Computer Science Department,
        IT University of Copenhagen, 2300 Copenhagen, Danmark
        {\tt\small mahmo@itu.dk, wasowski@itu.dk}}%
\thanks{$^{2}$Tobias John and Einar Broch Johnsen are with the Department of Informatics, University of Oslo,
        Oslo, Norway
        {\tt\small tobiajoh@ifi.uio.no, einarj@ifi.uio.no}}%
\thanks{$^{3}$Jeremy Coffelt is with Subsea R\&D group of Rosenxt,
    Bremen, Germany
    {\tt\small jcoffelt@rosen-nxt.com}}%
}

\makeatletter
\def\endthebibliography{%
  \def\@noitemerr{\@latex@warning{Empty `thebibliography' environment}}%
  \endlist
}
\makeatother

\begin{document}

\maketitle
\thispagestyle{empty}
\pagestyle{empty}

\begin{abstract}
  Autonomous Underwater Vehicles (AUVs) need to operate for days without human
  intervention and thus must be able to do efficient and reliable task
  planning. Unfortunately, efficient task planning requires deliberately
  abstract domain models (for scalability reasons), which in practice leads to
  plans that might be unreliable or under performing in practice.  An optimal
  abstract plan may turn out suboptimal or unreliable during physical
  execution. To overcome this, we introduce a method that first generates a
  selection of diverse high-level plans and then assesses them in a low-level
  simulation to select the optimal and most reliable candidate. We evaluate the
  method using a realistic underwater robot simulation, estimating the risk
  metrics for different scenarios, demonstrating feasibility and effectiveness
  of the approach.
\end{abstract}

\section{Introduction}

\noindent
Autonomous Underwater Vehicles (AUVs) support unmanned operations,
such as gathering data for scientific and commercial purposes.  The promise of
autonomy makes AUVs ideal for inspection tasks in harsh environments,
as operating in these conditions poses a high risk to human workers.
However the high cost of vehicles, operation in the vicinity of
high-value assets, and the risk of causing ecological disasters introduce
a requirement of reliable and safe autonomy\,\cite{chen2021review}.
\looseness -1

Traditionally, adaptive planning has been a key technology to achieve
autonomy.  Mainstream work on planning optimizes for expected rewards.
However, ignoring the probability of achieving the expected reward
(the \emph{uncertainty}) seems overly optimistic for high risk
operations. Although the expected reward might be high, an extremely
different outcome may have similar or higher probabilities than the
expectation.  \looseness -1


\begin{figure}[t!]
  \includegraphics [
    width = \linewidth,
    clip,
    trim = 0mm 15mm 0mm 34mm
  ] {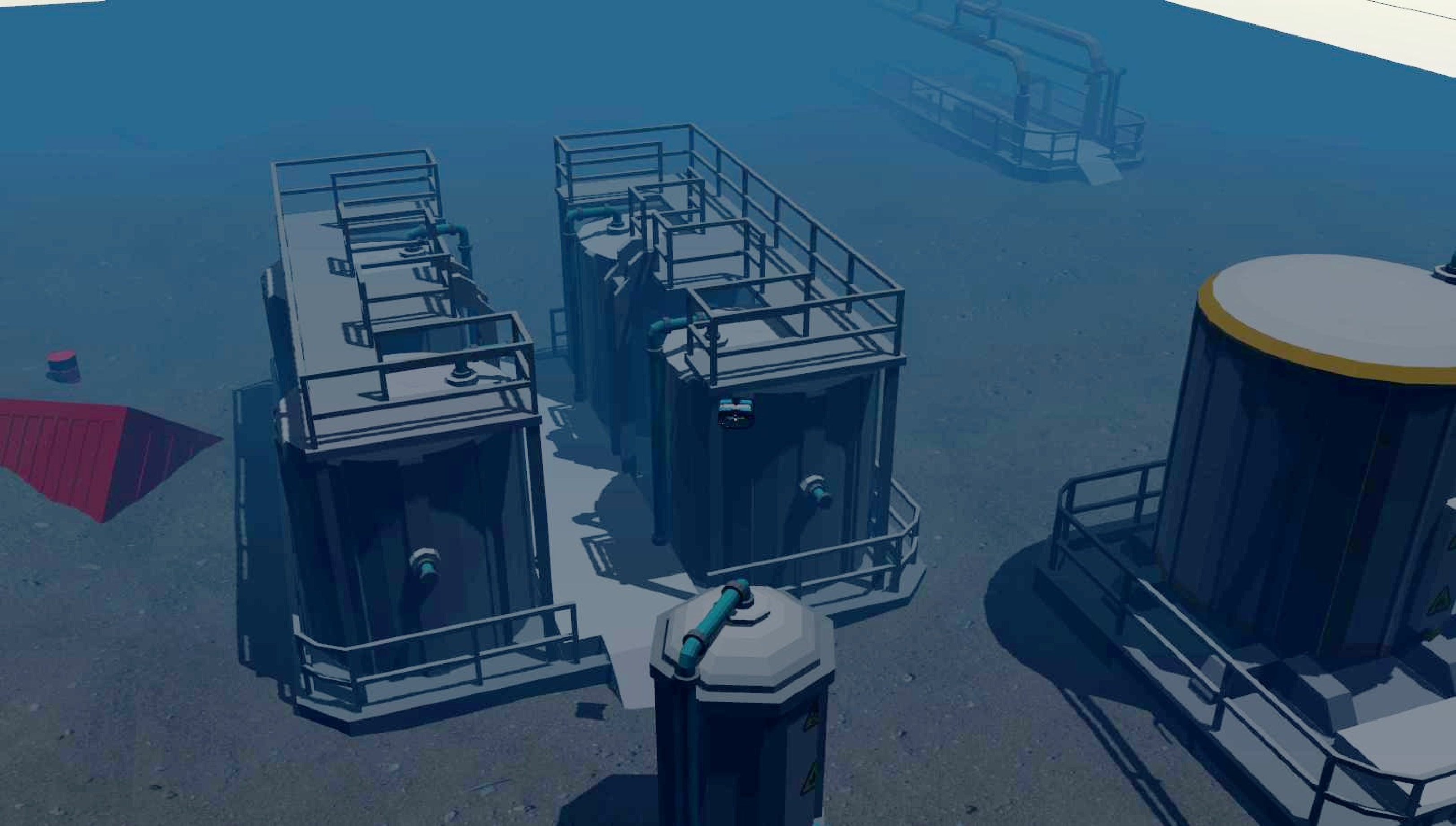}
  \caption{An underwater scene visualized in Gazebo}%
  \label{fig:sim_overview}
\end{figure}


\begin{figure}[t!]
  \resizebox{\linewidth}{!}{\input{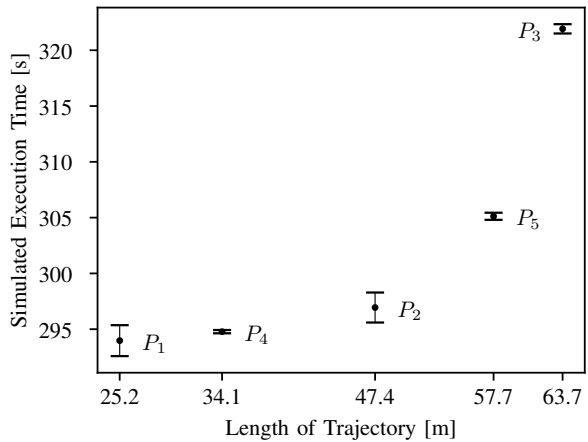}}
\caption{Comparison of simulated execution times for different plans. Plan IDs refer to those in \cref{table:plans}.}%
  \label{Fig:comparison}
    \vspace{-0.5cm}

\end{figure}


\begin{example}
  Consider a marine inspection scenario where the task is to safely
  inspect multiple sub-sea installations, e.g. the vertical small
  tanks in \cref{fig:sim_overview}. During the mission, an AUV
  needs to choose between five paths $P_1, \ldots, P_5$. A typical
  planner here finds that plan $P_1$ is the \emph{shortest} and
  selects it. If we simulate the different plans, we see that
  $P_1$ leads in general to a \emph{smaller mean execution time}
  than the other plans, but the \emph{variance of the execution time} is
  high, i.e., there is always the risk that the execution time is way
  higher than expected beforehand and possibly higher than the execution time of $P_4$. Arguably, $P_4$ might be the safest plan, with a decent
  expected execution time and a very small variance.
  \looseness -1

\end{example}

Two key problems with high-level planning lead to situations like in \cref{Fig:comparison}.  First, high-level planning deliberately operates under reduced information.  Second, the uncertainty of obtaining the reward is typically ignored.
Although handling uncertainty in
planning systems seems necessary in dynamic environments, most
existing research on underwater mission planning (e.g., on
re-planning\,\cite{hoffmann2001ff,DBLP:journals/corr/abs-1109-6051} or
temporal planning\,\cite{benton2012temporal}) defines planning
problems as optimization against the mean reward objective such as time or
energy\,\cite{cashmore2014auv}.

To address these problems, we propose a method that enables marine robots to
select reliable plans by assessing high-level plan candidates using a
low-level simulation model and statistical assessment of uncertainty. The goal
of this work is to increase the robustness of planning for autonomous
underwater robots in dilemma situations. We contribute a framework with the
following distinct features:
\begin{enumerate}[(i)]
\item Generation of probabilistic planning problems from sonar sensor
  data,

\item Transformation of planning problems to generate a selection of
  candidate plans that involve different degrees of risk-averseness,

\item Risk evaluation for candidate plans using realistic underwater
  simulation, and

\item Use of several risk metrics for selecting plans.

\end{enumerate}


\section{Related Work}

\noindent
We identified several lines of research on planning for AUVs that are related to our work: (A) modelling complex environments, (B) handling outcome uncertainties for various actions, (C) explaining generated plans and (D) risk-averse planning.\looseness -1

(A) The planning time depends significantly on how the planning domain
is formulated.  Steenstra shows how to compensate for environmental
challenges, such as location uncertainty and communication
limitations\,\cite{steenstra2019pddl}, using the Problem Domain
Definition Language (PDDL) to model an abstract-level planning problem
for Lightweight Autonomous Unmanned Vehicles\,\cite{sousa2012lauv}.  In
his work, a temporal deterministic planner
OPTIC\,\cite{benton2012temporal} consistently produces high-quality
plans with the lowest cost, however at a prohibitively high computation
cost. To overcome this, Steenstra translates the planning domain into
Event-B~\cite{fourati2016verification}, reducing the
planning time. We address this issue differently, by using a simpler
abstract planning model and statistical tests to select between several
best plans.\looseness -1

(B) According to Cashmore et al., the efficiency of long-horizon mission
planning is better and more reliable if the uncertainty is neglected in
favor of re-planning~\cite{cashmore2016opportunistic,cashmore2017opportunistic}. We
argue that neglecting uncertainty is not generally useful; re-planning
happens too late, after the robot has already committed to a (possibly)
unreliable plan. Replanning is a good adjustment strategy, complementary to what we propose---choosing a reliable plan to begin with.
\looseness -1



(C) 
Carreno et al. support
investigation of multiple planning choices at the planning and execution time,
although finding appropriate explainable metric(s) still needs to be
investigated~\cite{carreno2021explaining}. We are not concerned with
explanations of plans here, but with the robot itself assessing the plans using
more faithful models than the planner, and selecting between several options.\looseness -1


(D) There were prior attempts to assess generated policies using variance of cumulative reward\,\cite{sato2001td}. Unfortunately, the risk-sensitivity is not compatible with classical planning algorithms. Our approach of risk-averse planning is inspired by the work of Koening and Simmons\,\cite{KOENIGHowMakeReactive1994} that explore the spectrum of risk-sensitivity from risk-seeking to risk-averse. The MDP emphasizing risk sensitivity to maximize expected exponential utility aligns with a robust Markov Decision Process (MDP) aimed at maximizing the worst-case criterion~\cite{moldovan2012risk}. We bound the risk factor to a small interval to make the problem more compatible with Moldovan's result~\cite{moldovan2012risk}, and integrate risk-averse planning into a proof-of-concept implementation for a marine robot.
\looseness -1





\section{A Method for Risk-Averse Planning}%
\label{sec:method}

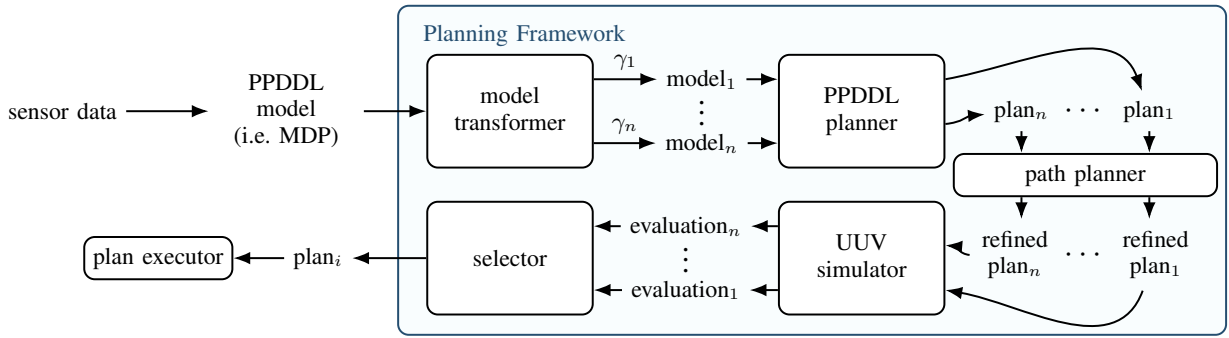
\begin{figure*}[t]
    \begin{tikzpicture} [
      scale = 0.85,
      every node/.style={font=\small}
    ]
    \draw (0,0) node (sensor) {sensor data};
		\draw ($(sensor) + (3.5, 0)$) node (model) {\begin{tabular}{c}
		     PPDDL \\
		     model \\
          (i.e.\ MDP)
		\end{tabular}};
		\draw ($(model) + (3.5, 0)$) node [bigCompontent] (MT) {\begin{tabular}{c}
		    model \\
		     transformer
		\end{tabular}};

        \draw ($(MT) + (3,0.5)$) node (model1) {model$_1$};
        \draw ($(MT) + (3,0.1)$) node {\large$\vdots$};
        \draw ($(MT) + (3,-0.5)$) node (modeln) {model$_n$};

        \draw ($(model1) + (2.5,-0.5)$) node [bigCompontent] (planner) {\begin{tabular}{c}
		     PPDDL \\
		    planner
		\end{tabular}};

        \draw ($(planner) + (2.5,0)$) node (plann) {plan$_n$};
        \draw ($(planner) + (3.5,0)$) node {\large$\cdots$};
        \draw ($(planner) + (4.5,0)$) node (plan1) {plan$_1$};

        \draw ($(plann) + (1, -1)$) node[wideComponent] (pathPlanner) {path planner};

        \draw ($(pathPlanner) + (-1.1,-1.25)$) node (rplann) {\begin{tabular}{c}
		     refined \\
		     plan$_n$
       \end{tabular}};
        \draw ($(pathPlanner) + (0,-1.25)$) node {\large$\cdots$};
        \draw ($(pathPlanner) + (1.1,-1.25)$) node (rplan1) {\begin{tabular}{c}
		     refined \\
		     plan$_1$
       \end{tabular}};

        \draw ($(planner) + (0,-2.3)$) node [bigCompontent] (Sim) {\begin{tabular}{c}
		     UUV \\
		     simulator
		\end{tabular}};

        \draw ($(Sim) + (-2.75,-0.5)$) node (evaluation1) {evaluation$_1$};
        \draw ($(Sim) + (-2.75,0.1)$) node {\large$\vdots$};
        \draw ($(Sim) + (-2.75,0.5)$) node (evaluationn) {evaluation$_n$};

        \draw ($(Sim) + (-5.5,0)$) node [bigCompontent] (selector) {selector};

        \draw ($(selector) + (-3,0)$) node (finalPlan) {plan$_i$};

        \draw ($(finalPlan) + (-2.5,0)$) node [component] (exec) {plan executor};

        \draw[operationFlow] (sensor) -- (model);
		\draw[operationFlow] (model) -- (MT);

        \draw[operationFlow] ($(MT.east) + (0,0.5)$) -- node[above] {$\uParam_1$} (model1);
        \draw[operationFlow] ($(MT.east) + (0, -0.5)$) -- node[above] {$\uParam_n$} (modeln);
        \draw[operationFlow] (model1) -- ($(planner.west) + (0, 0.5)$);
        \draw[operationFlow] (modeln) -- ($(planner.west) + (0, -0.5)$);

        \draw[operationFlow] ($(planner.east) + (0,0.5)$) to[out=10, in=110] (plan1);
        \draw[operationFlow] ($(planner.east) + (0,-0.2)$) to[out=5, in=185] (plann);

        \draw[operationFlow] (plan1) -- ($(pathPlanner.north) + (1,0)$);
        \draw[operationFlow] (plann) -- ($(pathPlanner.north) + (-1,0)$);

        \draw[operationFlow] ($(pathPlanner.south) + (-1,0)$) -- ($(rplann.north) + (0.1,0)$);
        \draw[operationFlow] ($(pathPlanner.south) + (1,0)$) -- ($(rplan1.north) + (-0.1,0)$);

        \draw[operationFlow] (rplan1) to[out=-110, in=-10] ($(Sim.east) + (0,-0.5)$);
        \draw[operationFlow] ($(rplann.west) + (0.2,0)$) to[out=-185, in=-5]($(Sim.east) + (0,0.2)$);

        \draw[operationFlow] ($(Sim.west) + (0,-0.5)$) -- (evaluation1);
        \draw[operationFlow] ($(Sim.west) + (0,0.5)$) -- (evaluationn);
        \draw[operationFlow] (evaluation1) -- ($(selector.east) + (0,-0.5)$);
        \draw[operationFlow] (evaluationn) -- ($(selector.east) + (0,0.5)$);

        \draw[operationFlow] (selector) -- (finalPlan);
        \draw[operationFlow] (finalPlan) -- (exec);

        \begin{pgfonlayer}{bg}
            \draw[component, background] ($(MT) + (-1.75, 1.65)$) rectangle ($(Sim) + (5.7, -1.2)$);
            \draw[text=mainblue] ($(MT) + (-1.5, 1.5)$) node[anchor=north west] {Planning Framework};

        \end{pgfonlayer}

	\end{tikzpicture}
  \vspace{-1mm}
  \caption{Flow diagram of proposed method.}%
  \label{fig:diagram}
  \vspace{-1mm}
\end{figure*}

\subsection{Framework}

\noindent
Both high-level task planning and low-level path planning are crucial for success of marine missions. A mission comprises a high-level plan and sub-plans generated by low-level (path) planners. Such low-level plans can be used for docking, for generating inspection trajectories around objects of interest (such as the fuel tanks shown in \cref{fig:sim_overview}), and for following waypoints selected by a robot operator.  The goal of high-level planning is to synthesize a plan to reach a state where the desired goals are achieved, given abstract descriptions of the initial state of the world and a set of possible actions. Symbolic AI planning is advantageous if the problem can be described declaratively and non-trivial domain knowledge exists and is relevant. However, symbolic planning necessarily, by-design, operates on highly approximate models of reality (abstractions), which introduces a significant performance gap with reality.  We address this by generating multiple abstract plans and evaluating their performance statistically at runtime, using a simulator, which has a much less abstract view of the world than the high-level planner.
\looseness -1


\cref{fig:diagram} summarizes the method. We assume that a probabilistic model of the current situation of the robot is available, extracted from sensor data, e.g.\ sonar data. This includes the possible actions of the robot and the goal of the mission, e.g.\ inspect the underwater infrastructure of interest. We generate different plans (with different sensitivity to risk) and use a (low-level) path planner to refine the high-level actions of the generated plans. Afterwards, we evaluate the performance of the refined plans using simulation. The generated evaluation for the different plans allows us to choose the plan that balances the involved uncertainty and the expected performance.
\looseness -1

\subsection{Plan Generation}

\noindent
We model high-level planning problems in \emph{Probabilistic Programming Domain Definition Language} (PPDDL), a representation of \emph{Markov Decision Processes} (MDPs)\,\cite{younes2004ppddl1}. An MDP is a
5-tuple \(\MDP = (S, A, \probMC, s_0, \rew)\) where $S$ is a finite
set of states, $A$ a finite set of
actions, $\probMC : S \times A \times S \rightarrow [0,1]$ the transition
function, $s_0 \in S$ the initial state, and $\rew : S \mapsto \mathbb{R}_{\geq 0}$ the reward.
A PPDDL problem specifies a set of goal states \(G \subseteq S\) that should be reached.

\begin{figure}
  \centering
    \resizebox{0.7\linewidth}{!}{
        \begin{tikzpicture}
            \draw (0,0) node[state] (initial) {initial};

            \draw ($(initial) + (5, 0)$) node[state] (oil) {\begin{tabular}{c}
                 \oilDrop  \\
                 tank
            \end{tabular}};

            \draw ($(initial) + (2.5, -2.6)$) node[state, accepting] (final) {\begin{tabular}{c}
                 \goalFlag  \\
                 final
            \end{tabular}};


            \draw[arrow] ($(initial.west) + (-0.5, 0)$) -- (initial);

            \draw[arrow, noRisk] (initial) -- node[mdpLabel, above] {no risk}  (final);
            \draw[arrow, noRisk] (final) to[bend left] node[mdpLabel, below] {no risk} (initial);

            \draw[arrow, noRisk] (final) -- node[mdpLabel, above] {no risk} (oil);
            \draw[arrow, risk] (oil) to[bend left=25] node[mdpLabel, below] {risk: collision} (final);

            \draw[arrow, risk] (initial) to[bend left=20] node[above] {risk: collision}(oil);
            \draw[arrow, noRisk] (oil) -- node[mdpLabel, below] {no risk} (initial);

        \end{tikzpicture}
    }
  \caption{Example scenario described by an MDP. It includes three waypoints, which are the critical ones for the mission.}%
  \label{fig:mdp}

  \vspace{-1mm}
\end{figure}
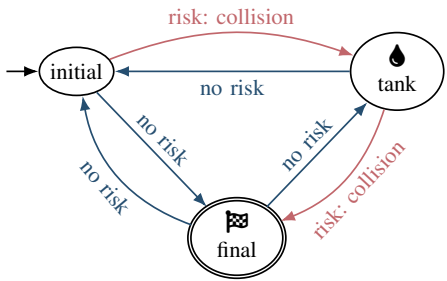

\begin{example}
    \cref{fig:mdp} shows an example of a planning model MDP. A robot can navigate between three waypoints and its mission is to inspect all tanks to make sure that there is no leakage. A non-deterministic choice (an action) determines which waypoint the robot inspects next. Some transitions involve risks, i.e.\ the AUV may collide with infrastructure with some probability. The goal is to reach the waypoint \enquote{final} after having performed an inspection at the waypoint ``tank.'' The cost is the length of the generated plan.  Note that this cost is deliberately abstract, not only for simplicity of the example, but to emphasize that the high-level planner always operates on an abstraction.
\end{example}

Given a plan \(\policy : S \mapsto A\), non-determinism in the MDP can be resolved to obtain the induced \emph{Markov Chain (MC)} $\MDP_{\policy} = (S, \probMC_{\policy}, s_0, \rew)$ where for each pair of states $s, s'$ the transition probability is defined by $\probMC_{\policy} (s, s') = \probMC (s, \policy(s), s')$. A \emph{history} $h$ is a finite sequence of states $\pathMC = (s_0, s_1, \ldots, s_n)$. Define a history's \emph{probability} by $\probMC (\pathMC) = \prod_{i=0}^{n-1} \probMC_{\policy}(s_i, s_{i+1})$ and its \emph{reward} by $\rew(\pathMC) = \sum_{i=0}^{n-1} \rew(s_i)$. For goal states $G \subseteq S$, the corresponding reward of the Markov Chain, $\rew_\MC (G)$, is a discrete random variable with:
\begin{equation}
  \Pr \left(\rew_\MC (G) \! = \! x \right) ~ =
  \hspace{-1em}
  \sum_{\pathMC = (s_0, \ldots, s_n)}
  \hspace{-1em}
  \probMC(\pathMC)
\end{equation}
where the summation ranges over all paths \(h\) such that \( s_n \in G\), \(s_0, \ldots, s_{n-1} \not\in G \), and  \(\rew(\pathMC) =  x\)




For a given MDP model, we generate a set of different \emph{candidate high-level plans}. Our method is agnostic to the algorithm to compute the different candidates, but in this paper, we use Koenig and Simmons' approach \cite{KOENIGHowMakeReactive1994} to generate risk-averse plans with a non-linear utility function for cumulative rewards.  This function values a difference between high rewards less than the same difference between smaller rewards; i.e., the cumulative reward of a few \enquote{best-case} paths has smaller impact on plan  selection than many paths with a decent cumulative reward.  The utility function depends on a parameter \uParam, which balances expected reward and involved risk. The main advantage of this construction is that we obtain a utility function for the cumulative reward for risk-sensitive planning by only making local changes; i.e., the probabilities of the MDP's transitions are replaced by values that take the probability,  immediate reward and the risk sensitivity into account.
\begin{definition}[Transformed Transition System]
    \label{transformedTS}
    Given an MDP $\MDP = (S, A, \probMC, s_0, R)$ and a parameter \(0 < \uParam < 1\), we define the transformed transition system $\MDP^{\uParam} =  (S, A, \probMC^{\uParam}, s_0)$ where  $\probMC^{\uParam} : S \times A \times S \rightarrow [-1, 0]$ and:
    \begin{equation}
      \probMC^{\uParam} (s, a, s') = \probMC(s,a,s') \cdot \left(-\uParam^{R(s)}\right)
    \end{equation}
\end{definition}%
Although the resulting transition system is not an MDP, standard algorithms can
be used to find the plan with maximal utility by maximizing the
\enquote{pseudo-probability} $\probMC^{\uParam}$ to reach a goal state\,\cite{KOENIGHowMakeReactive1994}.
The transformation described in \cref{transformedTS} can be done at the level of the actions in Probabilistic PDDL (PPDDL) domains.

We call a PPDDL planner for different random values of \uParam from the interval $(0,1)$ to generate the optimal plan for each value. Each optimal plan is simply the plan that maximizes the probability of reaching one of the goal states in the transformed model. For different values of \uParam, we get different plans, with different levels of risk-aversion. A value of $\uParam = 1$ means that the planner only tries to optimize expected cost, while a value of $\uParam = 0$ means that the planner tries to minimize the cost that can be guaranteed.  These plans form our set of candidate plans.

Crucially, plans are always generated with respect to a limited number of risk metrics, as planners can only handle a limited number of objectives at a time. Most often, as in our case, only one parameter  is considered (here \uParam).  This risk measurement might---but need not---correlate to other risk metrics, e.g., variance or entropy of the cumulative reward.  Therefore, we propose to separate the assessment of candidate plans against risk measurements of interest from the plan generation.
\looseness -1

\subsection{Plan Evaluation}

\noindent
To select the best plan, we use metrics based on the probability distribution of the reward $\estimatedDis = \rew_{\MDP_{\policy}}(G)$ to evaluate each of the candidate plans in the Bayesian spirit. One possibility is to use Monte Carlo simulation in the MDP of the planning problem. This, however, is typically insufficient, and definitely too unreliable for underwater robots. An MDP model for behavior abstracts away too much information about the physical environment. For example, our model neglects critical considerations such as unpredictable water currents, GPS-denied vehicle localization, or noisy acoustic sensors. In order to overcome this limitation, we perform the plan assessment not in the MDP, but in a more detailed model, using an \emph{underwater physics simulator} that provides increased realism. This involves using low-level planners to refine the actions in our generated high-level plan.  In order to get a good estimation of the risk associated with the plans, each generated plan needs to be simulated multiple times. The simulator uses random variables for environment conditions, e.g.\ underwater currents or water visibility, to obtain a representative sample of runs.
\looseness -1

We use different metrics to assess the current and anticipated risks for the above mathematical model.

\begin{enumerate}

    \item The \emph{expected reward} $\E{\estimatedDis}$ is typically a primary concern when selecting a plan. Even in risk-averse settings, a decent expected reward is required.
    \item The \emph{variance} of the reward $\E{\estimatedDis- \E{\estimatedDis}^2}$ is an often used measurement for risk~\cite{geibel2005risk}. The higher the variance, the more risk is involved.

    \item The \emph{entropy} of the reward $\textstyle -\sum_{x\in R} \Pr(\estimatedDis = x) \log_2 (\Pr(\estimatedDis = x))$ is another common metric used to measure the uncertainty or ``surprise'' of a reward. 

\end{enumerate}

\noindent
After simulating the candidate plans and estimating the risk metrics of interest, we select the plan that balances the different metrics best. Afterwards, we hand over the selected plan to the plan executor, which runs the plan on the actual robot.
\looseness -1



\section{Experimental Results}

\noindent
The goal of the experiment is to demonstrate that our framework can generate a reliable plan for an AUV given an inspection mission, even if this plan is not optimal with respect to the expected cost in the high-level planning model. We used a scenario where the underwater robot needs to inspect an area of interest while avoiding risky places. Our scenarios always allow for several different strategies or plans to accomplish the mission.
\looseness -1

\subsection{Evaluation Scenario}%
\label{subsec:design}






\begin{figure}[t]
 \includegraphics [
    width = \linewidth,
    clip,
    trim = 0mm 25mm 0mm 20mm
  ] {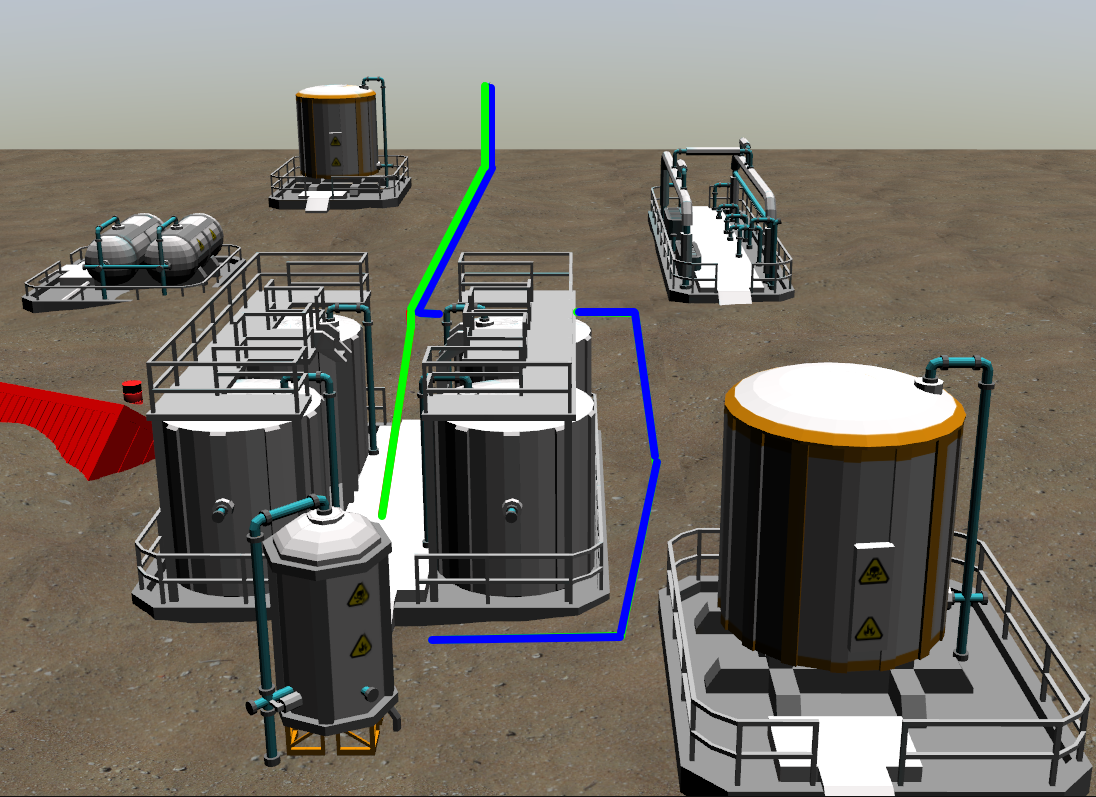}
  \caption{Blue line: a plan fragment following the safest path as returned by our method. Green line: a risky path suggested by a high-level risk-neutral planner}
\label{fig:RViz}
\end{figure}







\noindent
The robot starts somewhere in the distance on the top of \cref{fig:sim_overview}. The goal is to navigate and inspect the tank on the bottom of the illustration. \cref{fig:RViz} shows two plans that demonstrate the trade-off between expected cost and risk. We show the safest plan, as proposed by our method, with a blue line, and a shorter plan with a green line, which requires the robot to navigate through the narrow path between the quad tanks. The longer plan, where the marine robot navigates around the large vertical tank, is safer and thus preferred. To avoid collisions, the robot needs to move slowly when close to the tanks, and then accelerate when there is enough space to do so safely. Therefore, the safer path would not result in a large delay.

We use plan length and the risk of collision with the infrastructure as cost metrics (negative reward).  Navigating close to objects, such as navigating through a narrow gap between tanks as in \cref{fig:sim_overview}, can be part of a plan with low cost, but with high risk.



We created a PPDDL model of the scenario, fixing the goal of the mission and the actions that the AUV can perform. The actions not only involve moving from one waypoint to another but also inspecting objects at 50 waypoints using a helical trajectory around the object. The details of the actions are refined by low-level path planners after the high-level plan is synthesized. To keep the state space as small as possible, we do not use all waypoints for the high-level planner, as the low-level path planner does, but we only consider critical waypoints. Such points are the ones close to one of the fuel tanks or between two narrow infrastructure objects.

We use six different inspection points including in front of the \emph{large vertical tank}, inside of the \emph{platform}, in front of the \emph{tank pairs}, in front of and behind the \emph{quad tank}, next to the \emph{large vertical tank clone} and in front of the \emph{small vertical tank}. We select four waypoints as critical: (1) right of the small vertical tank, (2) in front of the narrow path in the quad tank, (3) between the quad tank and the largest tank and (4) left of the small vertical tank. The velocity of robot is decreased in vicinity of critical waypoints to avoid collisions.

To create risk-aware variants of the model, we sample risk factors $\uParam$ with a random uniform distribution over $[0.4,1)$ and transform the mode as in \cref{transformedTS}. This interval provides a broad spectrum of plans, while still ensuring that the plans optimize the expected reward sufficiently. We use an off-the-shelf PPDDL planner to generate plans for the generated transformed models. Afterwards, we refine the plans by instantiating the actions using low-level planners for path planning. After running the refined plans in Unmanned Underwater Vehicle (UUV) simulations, we estimate the metrics expected value, variance and entropy for the execution time. We select the plan based on these metrics.
\looseness -1

At runtime, during plan assessment in simulation, a collision detector monitors the distance between the robot and the infrastructure objects and logs incidents.
\looseness -1

\begin{figure}[t!]

  \vspace {-1mm}

  \caption{A seabed OctoMap\,\cite{hornung2013octomap}, using a low resolution (1000 samples with 0.5 resolution) sufficient for high-level planning. Each voxel represents a probability of occupancy.\label{fig:OctoMAp}}

  \includegraphics [
    width = \linewidth,
    clip,
    trim = 10mm 10mm 0mm 16mm
  ] {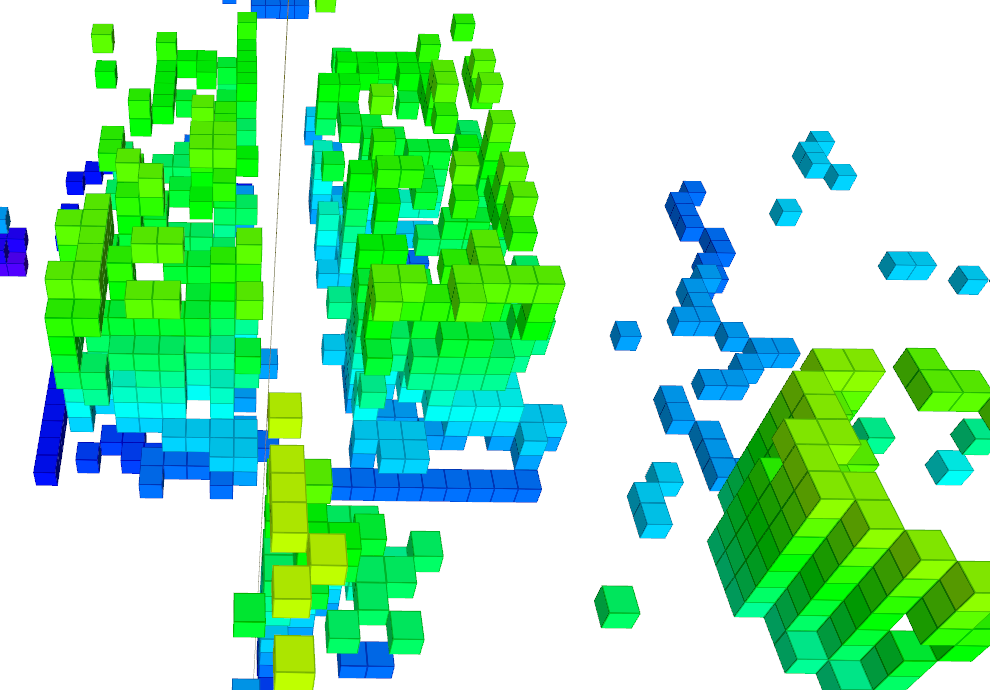}

\end{figure}

\vspace{0.00mm}

\begin {table*} [t]

  \renewcommand \arraystretch {1.02}
  \centering
  \resizebox{\linewidth}{!}{
  \begin{tabular}{rlrrr|rrrr}

  &
  & \textbf {planning}
  & \multicolumn{1}{r}{\textbf {high-level}}
  & \textbf {low-level}
  & \multicolumn 3c {\bfseries assessment (exec.\ time)}
  \\

    \textbf{ID}
  & \textbf{Plan schema}
  & \multicolumn{1}{r}{\textbf{time\,[s]}}
  & \multicolumn{1}{r}{\textbf{length}}
  & \multicolumn{1}{r}{\textbf{length}}
  & \multicolumn{1}{r}{\textbf{mean\,[s]}}
  & \multicolumn{1}{r}{\textbf{\llap{va}riance}}
  & \multicolumn{1}{r}{\textbf{\llap{en}tropy}}
  \\[1.5mm]

  \(P_1\)
  & lg tank $\rightarrow$ quad tank $\rightarrow$ sm tank
  & 0.09
  & 3
  & 25.23
  & 293
  & 1.92
  & 1.6
  \\

  \(P_2\)
  & lg tank $\rightarrow$ tank pairs $\rightarrow$ platform $\rightarrow$ quad tank $\rightarrow$ sm tank
  & 0.11
  & 5
  & 47.48
  & 297
  & 1.8
  & 1.6
  \\

  \(P_3\)
  & lg tank $\rightarrow$ tank pairs $\rightarrow$ lg tank back $\rightarrow$ platform $\rightarrow$ quad tank $\rightarrow$ sm tank
  & 0.15
  & 6
  & 63.73
  & 322
  & 0.17
  & 1.6
  \\

  \(P_4\)
  & lg tank $\rightarrow$ quad tank $\rightarrow$ lg tank clone $\rightarrow$ sm tank
  & 0.07
  & 4
  & 34.19
  & 294
  & 0.02
  & 0.2
  \\

  \(P_5\)
  & lg tank $\rightarrow$ tank pairs $\rightarrow$ lg tank back $\rightarrow$ quad tank $\rightarrow$ sm tank
  & 0.11
  & 5
  & 57.73
  & 305
  & 0.1
  & 1.6
  \\

  \end{tabular}
    }

  \caption{Generated plans using risk-averse planning. We have 5 types of infrastructure in the simulation. The names \enquote{lg tank} and \enquote{sm tank} stand for \enquote{large vertical tank} and ``small vertical tank.'' The high-level length is measured in the number of steps (plan depth), and the low-level length is the total distance between all waypoints in meters. The vertical bar separates the high-level and low-level planning from plan assessment. The IDs are the same as in \cref{Fig:comparison}}%
  \label{table:plans}

  \vspace {-.5mm}

\end{table*}

\subsection{Design and Implementation}

\noindent
We execute the experiments on a laptop with 2.60GHz Intel Core CPU (i7-10750H) and 32GB of RAM, running Ubuntu 20.04. The implementation is based on ROS1 Noetic, Gazebo\,11, and a modified version of UUV simulator.\footnote{\url{https://github.com/mahyamkashani/uuv\_simulator}} The source code to reproduce our experiments is available online.\footnote{\url{https://github.com/remaro-network/risk-averse\_planning}}

We created the scenarios for gas and oil infrastructure inspection by using public assets from the DAVE project.\footnote{\url{https://field-robotics-lab.github.io/dave.doc/}} The 3D models can be obtained from the TurboSquid website.\footnote{\url{https://www.turbosquid.com/3d-models/3d-fuel-tank-1443266}}
\looseness -1

We estimate the occupancy probability for the discretized (high-level) map using OctoMap \cite{hornung2013octomap},
calculating the probability that a section of the environment is occupied using
simulated forward multibeam p900 sonar data.\footnote{\url{https://github.com/Bluerov2}} We tried this procedure with different resolutions starting from 0.05 up to 0.5 resolution. We settled on the latter (low-resolution), judging that this is suitable for high-level task abstraction.  \cref{fig:OctoMAp} shows an example octomap.  Each color represent a probability between 0 and 1. The values of the cubes (voxels) are used to create the state space of our planning problem.




We use a modified version of safe-planner to generate different risk-averse plans\,\cite{mokhtari2021safe}. The modifications comprise  incorporating the risk parameter \uParam. The output for each plan generation is a JSON file containing the plan and a statistics file that documents the performance of the produced plan. LIPBInterpolator (Linear Parabolic Interpolator) is used as a low-level path planner,\footnote{\url{https://uuvsimulator.github.io/packages/uuv_simulator/docs/python_api/uuv_trajectory_generator/lipbinterpolator}} to refine the actions of the high-level plans with intermediate waypoints.

We execute each resulting plan 10 times to evaluate its performance under slight randomization. We perturb positions of the vertical small tank in the simulation environment with random noise to mimic uncertainty in a physical environment, and use it to estimate the risk metrics of the plan (expected reward, variance, and entropy of the execution time).

\subsection{Results}

\noindent
\cref{table:plans} summarizes the generated plans and sub-plans for one of our evaluation scenarios (scenario ID 3 with five critical states in \cref{table:scenarios}). Interestingly, the variation in length at the low-level is lower than at the high-level.  While the former only differ by about 10\%, the latter can be 100\% longer, see for example $P_1$ and $P_3$.  This illustrates how typically, and deliberately, abstraction hides information from high-level models; underlining that a high-level planner always operates under imperfect information.   This is an inherent problem. In practice it is unrealistic to expect that the high-level model will be a sound abstraction from the perspective of all metrics of interest. This also indicates that it might be useful to consider several reasonable plans and evaluate them at lower level of abstraction. (Note that optimal planning at the low-level of abstraction is too expensive.)
\looseness -1


While the length of the low-level plans does not differ much, the simulation-based assessment shows that the actual execution times still can differ significantly (cf.\ \cref{Fig:comparison} and the mean time in \cref{table:plans}).  Plan \(P_1\), the shortest in the abstract domain, turns out to be relatively slow in a more detailed simulation with the real robot controller. The high variance also indicates unreliable execution. It seems more desirable to execute plan \(P_4\) in the physical environment, since this plan has both low and predictable running time.  In extreme cases, not observed in this experiment, the best plan according to the mean reward might be unusable, because its high variance makes it less desirable than another plan with worse mean but smaller variance.  In our scenario, the entropy is the same for all the candidate plans.  This is however not always the case.  Entropy is a more useful assessment metric in scenarios when distribution of rewards is multimodal.
\looseness -1

\begin{table}[b!]

  \renewcommand \arraystretch {1.05}

  \caption{Different scenarios for risk-averse planning, between 109 and 122 way points. All scenarios are solvable and the method produces the safest plan.}%
  \label{table:scenarios}


  \vspace {-.5mm}

  \hfill\begin{tabular}{rrr|rrr}

  & \multicolumn{2}{c}{\bfseries Problem properties}
  & \multicolumn{3}{c}{\bfseries Properties of the safest plan}
  \\[1.0mm]

    \multicolumn{1}{r}{\textbf{ID}}
  & \multicolumn{1}{r}{\textbf{\llap{d}epth\,[state]}}
  & \multicolumn{1}{r}{\textbf{\llap{\#c}rit.\,state\rlap{s}}}
  & \multicolumn{1}{r}{\textbf{\llap{l}ength}}
  & \multicolumn{1}{r}{\textbf{\llap{r}isk\,(\uParam)}}
  & \multicolumn{1}{r}{\textbf{\llap{pl}anning\,time\,[s]}}
  \\[2.0mm]

  1 & 3  & 3  & 5 & 0.95 & 0.05
  \\

  2 & 4  & 4  & 5 & 0.74 & 0.06
  \\

  3 & 5  & 5  & 6 & 0.40 & 0.11
  \\

  4 & 6  & 6  & 17 & 0.57 & 0.10
  \\[2mm]

  5 & 10 & 5  & 13 & 0.67 & 0.11
  \\

  6 & 15 & 5  & 13 & 0.49 & 0.14
  \\

  7 & 20  & 6 & 17  & 0.84 & 0.08
  \\

  8 & 30  & 10 & 52  & 0.58 & 0.13
  \\[2mm]

  9 & 40  & 15 & 66  & 0.84 & 0.16
  \\

  10 & 50  & 15 & 86  & 0.96 & 0.21
  \\

  11 & 60  & 15 & 106  & 0.97 & 0.29
  \\

  12 & 70  & 15 & 126  & 0.95 & 0.40
  \\[2mm]

  13 & 80  & 15 & 146  & 0.59 & 0.54
  \\

  14 & 90  & 15 & 166  & 0.57 & 0.93
  \\

  15 & 90  & 25 & 166  & 0.42 & 0.98
  \\

  16 & 90  & 35 & 166  & 0.75 & 1.04

  \end{tabular}\hfill\strut
\end{table}

\cref{table:scenarios} summarizes properties of optimal plans generated for different evaluation scenarios. The scenarios differ in the number of critical states in the high-level planning problem; and increase in complexity with the number of critical states.

We find that the method scales---all scenarios are solvable; i.e. we can generate a plan to reach the goal of the mission.
In all scenarios, the path planner generates between 109 and 129 waypoints, depending on the plan length. The planning time and the length of the found plan increase with the number of critical states as expected---the planner needs to search through a larger state space. The crucial aspect of this method is that when the planner stops scaling, one can abstract the domain model more aggressively (to make it scale), and use simulation-based assessment to select reasonable plans.  Notably, the risk parameter is often smaller than one. This indicates, that the found plan, which is the safest for the scenario, is not the one that minimizes cost, i.e. plan length, but one finds a trade-off between cost and risk aversion. This shows that the method is making non-trivial choices, comparing to a planner that only optimizes mean cost.
\looseness -1

We have experimented with increasing branching in the planning model.
Branching is typical for ordered tasks like manipulation. Since our plan generation is based on Breadth-First Search, the planning time does increase exponentially. This can be improved by switching to a different plan generation method, while keeping the same simulation-based plan assessment. In pipeline inspection tasks, which is the main use case in our project, we do not experience high branching.
\looseness -1




\section{Conclusion and Outlook}

\noindent
We have presented a risk-averse planning method based on generating multiple plans and separately, performing their simulation-based assessment. We implemented and evaluated the method, showing that it scales for simple inspection tasks in the underwater domain.  The evaluation has demonstrated the effectiveness of generating diverse plans, the effectiveness of simulation-based assessment in reducing the information-gap between high-level and low-level dynamics, and the effectiveness of the method overall to select the safest plan from a small list of candidates.

Notably, the separation of planning from assessment is applicable to different plan generation methods, not necessary ours. It can be used with a diversity of risk and performance metrics, such as (i) \emph{reward-bounded probability}, the probability that the reward falls below a given bound, designed to estimate how often bad runs occur, (ii) \emph{value at risk}, the maximum reward expected given a time horizon and a confidence level, capturing tail risk, and (iii) \emph{expected shortfall}, the average of the worst percentile of losses. It is not required that the high-level planning and the low-level assessment use the same performance objective.  One can, for instance, plan for minimal distance traveled, but assess against the simulated power consumption.  In the future, we intend to exploit these flexibility in planning underwater inspection scenarios.
\looseness -1




\bibliographystyle{IEEEtran}
\bibliography{main}


\end{document}